\documentclass{article}

\usepackage{graphicx,amscd,amsmath,amssymb,verbatim}
\usepackage[dvips]{hyperref}
\usepackage[TS1,OT1,T1]{fontenc}

\begin{document}

\title{Invariant Stochastic Encoders\footnote{Full version of a
short paper that was published in the Digest of the 5th IMA International
Conference on Mathematics in Signal Processing, 18-20 December 2000,
Warwick University, UK.}}
\author{Stephen Luttrell}
\maketitle

\noindent {\bfseries Abstract:} The theory of stochastic vector
quantisers (SVQ) has been extended to allow the quantiser to develop
invariances, so that only "large" degrees of freedom in the input
vector are represented in the code. This has been applied to the
problem of encoding data vectors which are a superposition of a
"large" jammer and a "small" signal, so that only the jammer is
represented in the code. This allows the jammer to be subtracted
from the total input vector (i.e. the jammer is nulled), leaving
a residual that contains only the underlying signal. The main advantage
of this approach to jammer nulling is that little prior knowledge
of the jammer is assumed, because these properties are automatically
discovered by the SVQ as it is trained on examples of input vectors.
\section{Introduction}

In vector quantisation a code book is used to encode each input
vector as a corresponding code index, which is then decoded (again,
using the codebook) to produce an approximate reconstruction of
the original input vector \cite{{Gray1984, GershoGray1992}}. The
standard approach to vector quantiser (VQ) design \cite{LindeBuzoGray1980}
may be generalised \cite{Luttrell1997} so that each input vector
is encoded as a {\itshape vector} of code indices that are stochastically
sampled from a probability distribution that depends on the input
vector, rather than as a {\itshape single} code index that is the
deterministic outcome of finding which entry in a code book is closest
to the input vector. This will be called a stochastic VQ (SVQ),
and it includes the standard VQ as a special case.

One advantage of using the stochastic approach is that it automates
the process of splitting high-dimensional input vectors into low-dimensional
blocks before encoding them, because minimising the mean Euclidean
reconstruction error can encourage different stochastically sampled
code indices to become associated with different input subspaces
\cite{{Luttrell1997b, Luttrell1999a}}. Another advantage is that
it is very easy to connect SVQs together, by using the vector of
code index probabilities computed by one SVQ as the input vector
to another SVQ \cite{Luttrell1999b}.

SVQ theory will be extended to the case of encoding noisy (or distorted)
data, with the intention of subsequently reconstructing an approximation
to the noiseless data. This theory is then applied to the problem
of encoding data vectors which are a superposition of a "large"
jammer and a "small" signal, where the signal is regarded as a distortion
superimposed on the jammer, rather than the other way around. The
reconstruction is then an approximation to the jammer, which can
thus be subtracted from the original data to reveal the underlying
signal of interest.

In Section \ref{XRef-Section-821212359} the underlying theory of
SVQs is developed together with its extension to the encoding of
noisy data, and in Section \ref{XRef-Section-82121246} some simulations
illustrating the application of SVQs to the nulling of jammers are
presented.
\section{Stochastic Vector Quantiser Theory}\label{XRef-Section-821212359}

In Section \ref{XRef-Section-82121255} the basic theory of folded
Markov chains (FMC) is given \cite{Luttrell1994}, in Section \ref{XRef-Section-821212540}
FMC theory is extended to the case of encoding noisy or distorted
data with the intention of eventually recovering the undistorted
data, in Section \ref{XRef-Section-821212555} this extended theory
is applied to the problem of encoding data that contain unwanted
"nuisance degrees of freedom", in Section \ref{XRef-Section-82121269}
some constraints (including the threshold trick of \cite{Webber1994})
on the optimisation of the encoder are introduced to encourage the
encoder to disregard the nuisance degrees of freedom (i.e. discover
invariances), and finally in Section \ref{XRef-Section-821212651}
this invariant encoder theory is applied to the problem of encoding
and subsequently nulling "large" jammers that obscure "small" signals.
\subsection{Folded Markov Chains}\label{XRef-Section-82121255}

The basic building block of the SVQ used in this paper is the folded
Markov chain (FMC) \cite{Luttrell1994}. An input vector $x$ is encoded
as a code index vector $y$, which is then subsequently decoded as
a reconstruction $x^{\prime }$ of the input vector. Both the encoding
and decoding operations are allowed to be probabilistic, in the
sense that $y$ is a sample drawn from $\Pr ( y|x) $, and $x^{\prime
}$ is a sample drawn from $\Pr ( x^{\prime }|y) $, where $\Pr (
y|x) $ and $\Pr ( x^{\prime }|y) $ are Bayes' inverses of each other,
as given by $\Pr ( x^{\prime }|y) =\frac{\Pr ( y|x) \Pr ( x) }{\int
dz \Pr ( y|z) \Pr ( z) }$, and $\Pr ( x) $ is the prior probability
from which $x$ is sampled.
\begin{figure}[h]
\begin{center}
\includegraphics{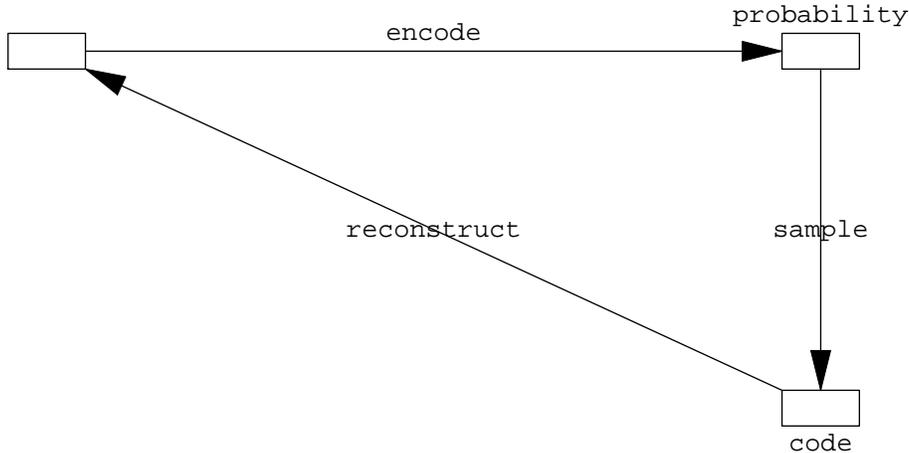}

\end{center}
\caption{A folded Markov chain (FMC) in which an input vector $x$
is encoded as a code index vector $y$ that is drawn from a conditional
probability $\Pr ( y|x) $, which is then decoded as a reconstruction
vector $x^{\prime }$ drawn from the Bayes' inverse conditional probability
$\Pr ( x^{\prime }|y) $.}
\end{figure}

In order to ensure that the FMC encodes the input vector optimally,
a measure of the reconstruction error must be minimised. There are
many possible ways to define this measure, but one that is consistent
with many previous results, and which also leads to many new results,
is the mean Euclidean reconstruction error measure $D$, which is
defined as \cite{Luttrell1994}
\begin{equation}
D\equiv \int dx \Pr ( x) \sum \limits_{y_{1}=1}^{M}\sum \limits_{y_{2}=1}^{M}\cdots
\sum \limits_{y_{n}=1}^{M}\Pr ( y|x) \int dx^{\prime }\Pr ( x^{\prime
}|y) \left| \left| x-x^{\prime }\right| \right| ^{2}%
\label{XRef-Equation-821232521}
\end{equation}

\noindent where $y=(y_{1},y_{2},\cdots ,y_{n}), 1\leq y_{i}\leq
M$ is assumed, $\Pr ( x) \Pr ( y|x) \Pr ( x^{\prime }|y) $ is the
joint probability that the FMC has state $(x,y,x^{\prime })$, $||x-x^{\prime
}||^{2}$ is the Euclidean reconstruction error, and $\int dx \sum
\limits_{y_{1}=1}^{M}\sum \limits_{y_{2}=1}^{M}\cdots \sum \limits_{y_{n}=1}^{M}\int
dx^{\prime }( \cdots ) $ sums over all possible states of the FMC
(weighted by the joint probability).

The Bayes' inverse probability $\Pr ( x^{\prime }|y) $ may be integrated
out of this expression for $D$ to yield \cite{Luttrell1994}
\begin{equation}
D=2\int dx \Pr ( x) \sum \limits_{y_{1}=1}^{M}\sum \limits_{y_{2}=1}^{M}\cdots
\sum \limits_{y_{n}=1}^{M}\Pr ( y|x) \left| \left| x-x^{\prime }(
y) \right| \right| ^{2}%
\label{XRef-Equation-821213533}
\end{equation}

\noindent where the reconstruction vector $x^{\prime }( y) $ is
defined as $x^{\prime }( y) \equiv \int dx \Pr ( x|y) x$. Because
of the quadratic form of the objective function, it turns out that
$x^{\prime }( y) $ may be treated as a free parameter whose optimum
value (i.e. the solution of $\frac{\partial D}{\partial x^{\prime
}( y) }=0$) is $\int dx \Pr ( x|y) x$, as required.
\subsection{Noisy Data}\label{XRef-Section-821212540}

\begin{figure}[h]
\begin{center}
\includegraphics{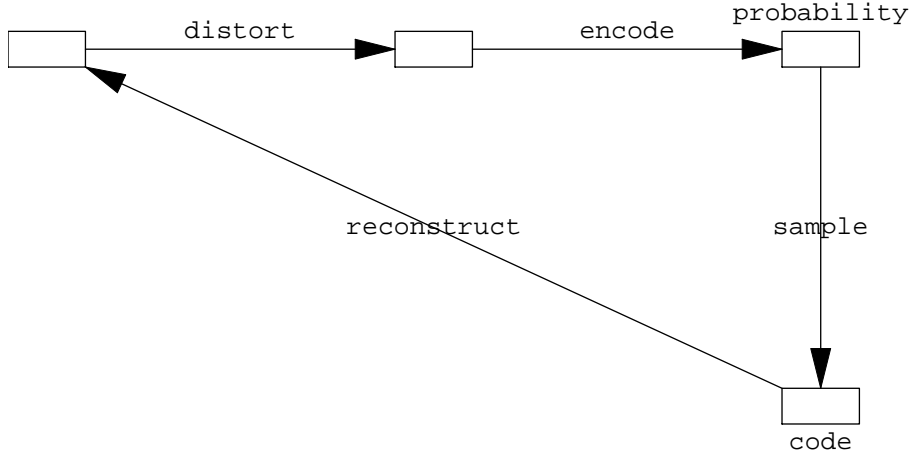}

\end{center}
\caption{A folded Markov chain (FMC) in which an input vector $x_{0}$
is first distorted into $x$, which is then encoded as a code index
vector $y$ that is drawn from a conditional probability $\Pr ( y|x)
$, which is then decoded as a reconstruction vector $x_{0}^{\prime
}$ drawn from the Bayes' inverse conditional probability $\Pr (
x_{0}^{\prime }|y) $.}\label{XRef-Figure-821213413}
\end{figure}

The FMC approach can be generalised to the problem of encoding noisy
or distorted data, with the intention of eventually recovering the
undistorted data. This generalisation is based on the results reported
in \cite{EphraimGray1988}. The input vector is $x_{0}$, which is
converted into the distorted input vector $x$ by a distortion process
$\Pr ( x|x_{0}) $, which is then encoded as a code index vector
$y$, which is then subsequently decoded as a reconstruction $x_{0}^{\prime
}$ of the original input vector. This is described by the directed
graph $x_{0}\longrightarrow x\longrightarrow y\longrightarrow x_{0}^{\prime
}$. The operations that occur are summarised in Figure \ref{XRef-Figure-821213413}.

The mean Euclidean reconstruction error measure $D$ becomes (compare
Equation \ref{XRef-Equation-821232521})
\begin{equation}
D=\int dx_{0} \Pr ( x_{0}) \int dx \Pr ( x|x_{0}) \sum \limits_{y_{1}=1}^{M}\sum
\limits_{y_{2}=1}^{M}\cdots \sum \limits_{y_{n}=1}^{M}\Pr ( y|x)
\int dx^{\prime }\Pr ( x_{0}^{\prime }|y) \left| \left| x_{0}-x_{0}^{\prime
}\right| \right| ^{2}
\end{equation}

\noindent The Bayes' inverse probability $\Pr ( x_{0}^{\prime }|y)
$ may be integrated out of this expression for $D$ to yield (compare
Equation \ref{XRef-Equation-821213533})
\begin{equation}
D=2\int dx_{0} \Pr ( x_{0}) \int dx \Pr ( x|x_{0}) \sum \limits_{y_{1}=1}^{M}\sum
\limits_{y_{2}=1}^{M}\cdots \sum \limits_{y_{n}=1}^{M}\Pr ( y|x)
\left| \left| x_{0}-x_{0}^{\prime }( y) \right| \right| ^{2}%
\label{XRef-Equation-821213948}
\end{equation}

\noindent where the reconstruction vector $x_{0}^{\prime }( y) $
is defined as $x_{0}^{\prime }( y) \equiv \int dx_{0} \Pr ( x_{0}|y)
x_{0}$, which may be treated as a free parameter.

Bayes' theorem $\Pr ( x_{0}) \Pr ( x|x_{0}) =\Pr ( x) \Pr ( x_{0}|x)
$ may be used to integrate out $x_{0}$ to yield
\begin{equation}
D=2\int dx \Pr ( x) \sum \limits_{y_{1}=1}^{M}\sum \limits_{y_{2}=1}^{M}\cdots
\sum \limits_{y_{n}=1}^{M}\Pr ( y|x) \left| \left| x_{0}( x) -x_{0}^{\prime
}( y) \right| \right| ^{2}+\operatorname{constant}%
\label{XRef-Equation-821222147}
\end{equation}

\noindent where $x_{0}( x) $ is defined as $x_{0}( x) \equiv \int
dx_{0} \Pr ( x_{0}|x) x_{0}$.

It is much more difficult to optimise this version of the objective
function than the version in Equation \ref{XRef-Equation-821213533},
because the $x_{0}( x) $ term is in general a non-linear function
of $x$. Worse still, the expression for $x_{0}( x) $ involves $\Pr
( x_{0}|x) $, which depends on the unknown $\Pr ( x_{0}) $, so $x_{0}(
x) $ cannot be computed analytically anyway. The situation looks
irretrievable, but it turns out that some progress can be made by
conceptually splitting $x$ into "signal" and "noise" subspaces,
as will be shown in Section \ref{XRef-Section-821212555}.
\subsection{Nuisance Degrees of Freedom}\label{XRef-Section-821212555}

\begin{figure}[h]
\begin{center}
\includegraphics{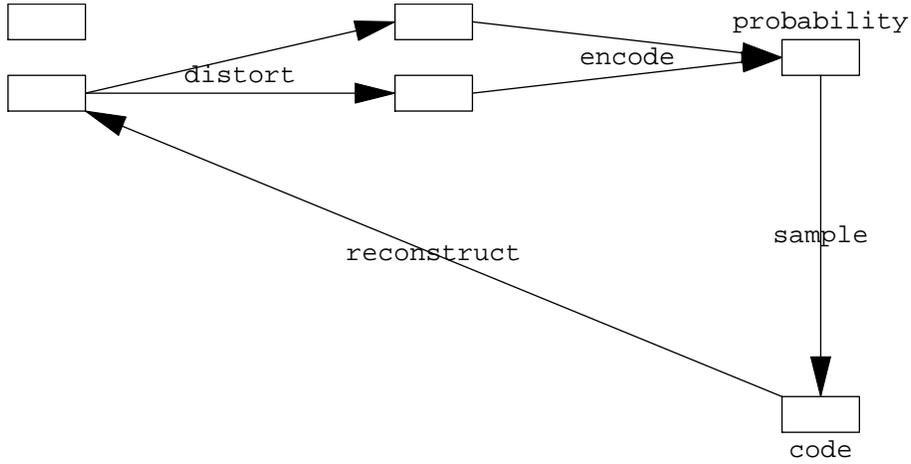}

\end{center}
\caption{A folded Markov chain (FMC) in which an input vector $(x_{0},0)$
is first distorted into $(x_{0},x_{\perp })$, which is then encoded
as a code index vector $y$ that is drawn from a conditional probability
$\Pr ( y|x_{0},x_{\perp }) $, which is then decoded as a reconstruction
vector $x_{0}^{\prime }$ drawn from the Bayes' inverse conditional
probability $\Pr ( x_{0}^{\prime }|y) $.}\label{XRef-Figure-82121399}
\end{figure}

For convenience, split up the input space into (possibly non-orthogonal)
subspaces as $(x_{0},x_{\perp })$, where all of the distortion is
contained in $x_{\perp }$, which requires that any distortion that
lies {\itshape in} the $x_{0}$ subspace is regarded as part of the
undistorted input. The directed graph becomes $(x_{0},0)\longrightarrow
\text{$(x_{0},x_{\perp })$}\longrightarrow y\longrightarrow x_{0}^{\prime
}$ as shown in Figure \ref{XRef-Figure-82121399}.

The expression for $D$ becomes (compare Equation \ref{XRef-Equation-821213948}).
\begin{equation}
D=2\int dx_{0}\Pr ( x_{0}) \int dx_{\perp } \Pr ( x_{\perp }|x_{0})
\sum \limits_{y_{1}=1}^{M}\sum \limits_{y_{2}=1}^{M}\cdots \sum
\limits_{y_{n}=1}^{M}\Pr ( y|x_{0},x_{\perp }) \left| \left| x_{0}-x_{0}^{\prime
}( y) \right| \right| ^{2}%
\label{XRef-Equation-821214120}
\end{equation}

\begin{figure}[h]
\begin{center}
\includegraphics{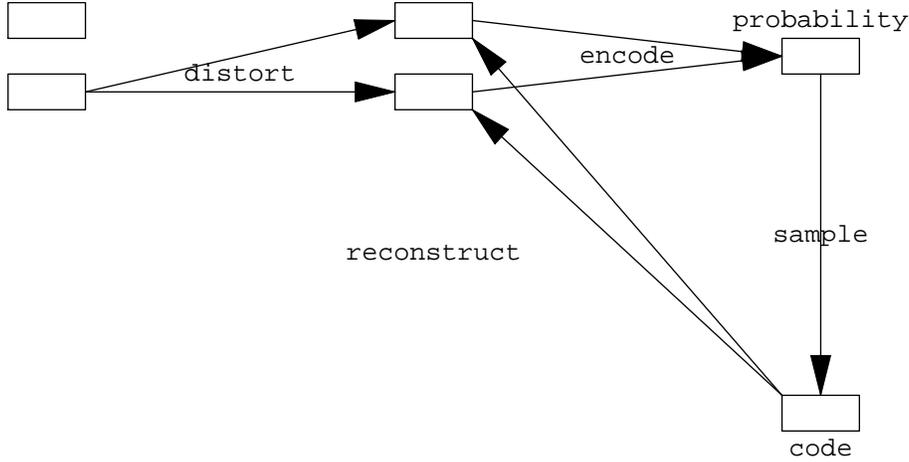}

\end{center}
\caption{Modified version of Figure \ref{XRef-Figure-82121399} in
which the reconstruction link is switched from the original undistorted
signal to the full signal+distortion.}\label{XRef-Figure-821214043}
\end{figure}

\noindent Consider the related optimisation problem in which an
attempt to to reconstruct $(x_{0},x_{\perp })$ is made, as shown
in Figure \ref{XRef-Figure-821214043}.
The corresponding objective function may be obtained by modifying
Equation \ref{XRef-Equation-821214120}, where the cross-term arising
from non-orthogonal $(x_{0},x_{\perp })$ is omitted.
\begin{equation}
\begin{array}{rl}
 D & =2\int dx_{0}\Pr ( x_{0}) \int dx_{\perp } \Pr ( x_{\perp }|x_{0})
\sum \limits_{y_{1}=1}^{M}\sum \limits_{y_{2}=1}^{M}\cdots \sum
\limits_{y_{n}=1}^{M}\Pr ( y|x_{0},x_{\perp })  \\
  &  \begin{array}{cc}
   &  
\end{array}\times \left( \left| \left| x_{0}-x_{0}^{\prime }( y)
\right| \right| ^{2}+\left| \left| x_{\perp }-x_{\perp }^{\prime
}( y) \right| \right| ^{2}\right) 
\end{array}%
\label{XRef-Equation-821214911}
\end{equation}

\begin{figure}[h]
\begin{center}
\includegraphics{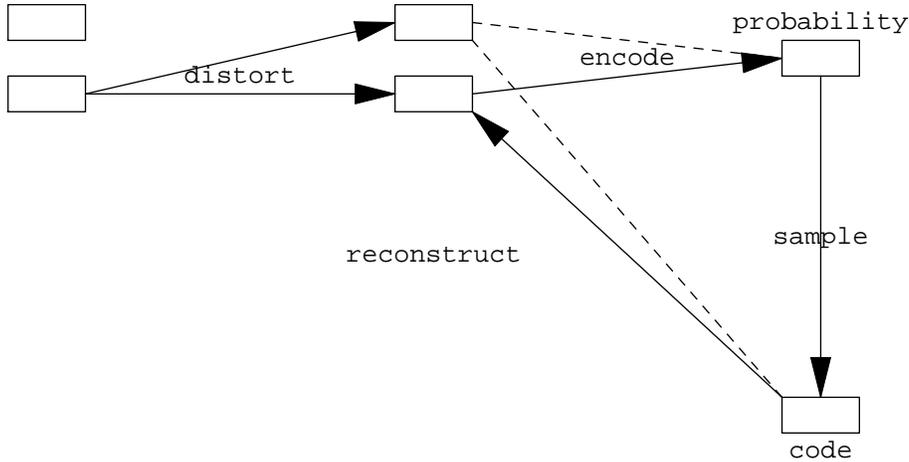}

\end{center}
\caption{Modified version of Figure \ref{XRef-Figure-821214043}
in which the encoder (and reconstruction) links from (and to) the
distortion subspace are deleted (as indicated by the dashed lines).}\label{XRef-Figure-821214223}
\end{figure}

\begin{figure}[h]
\begin{center}
\includegraphics{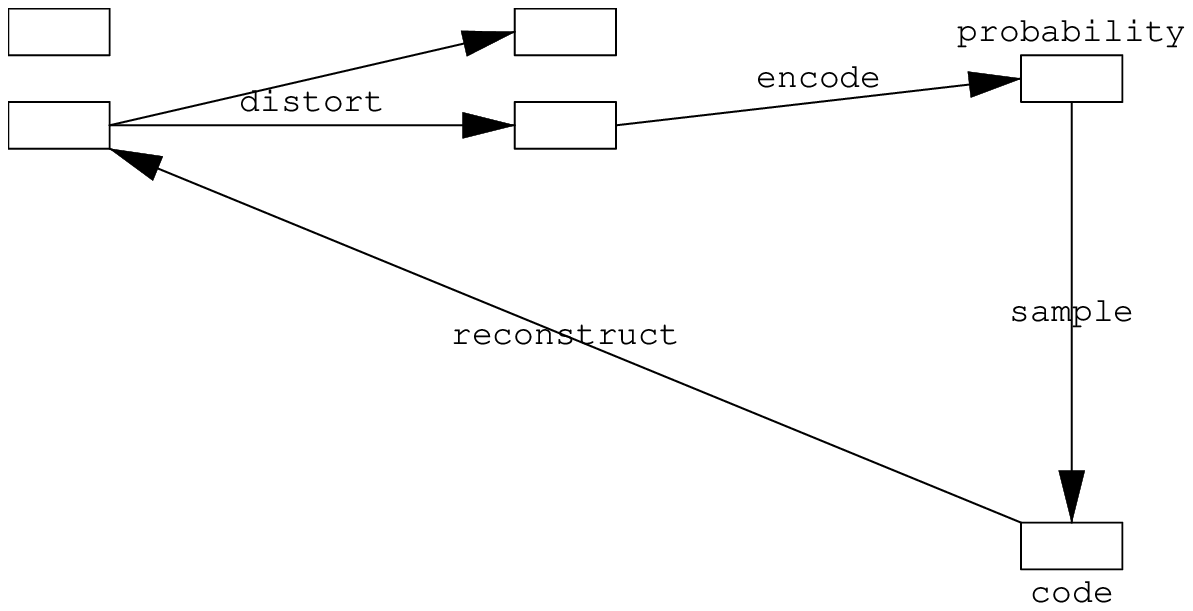}

\end{center}
\caption{Equivalent version of Figure \ref{XRef-Figure-821214223}
in which the reconstruction link is moved to an equivalent position.}\label{XRef-Figure-821214420}
\end{figure}

\noindent Assume for now (to be justified below) that some of the
links in Figure \ref{XRef-Figure-821214043} are broken as shown
in Figure \ref{XRef-Figure-821214223}.
Because the distortion subspace is not involved in the computations
in Figure \ref{XRef-Figure-821214223}, it may be redrawn as shown
in Figure \ref{XRef-Figure-821214420}.
This is the same as Figure \ref{XRef-Figure-82121399}, except that
the encoder now disregards (or is invariant with respect to) the
nuisance degrees of freedom.

In order to break the links as shown in Figure \ref{XRef-Figure-821214223}
the following argument is required:
\begin{enumerate}
\item Assume that the encoder is independent of $x_{\perp }$, so
that $\Pr ( y|x_{0},x_{\perp }) =\Pr ( y|x_{0}) $.
\item The $||x_{\perp }-x_{\perp }^{\prime }( y) ||^{2}$ term in
$D$ needs to simplify to a constant.
\item This requires that $\int dx_{0}\Pr ( x_{0}) \int dx_{\perp
} \Pr ( x_{\perp }|x_{0}) \sum \limits_{y_{1}=1}^{M}\sum \limits_{y_{2}=1}^{M}\cdots
\sum \limits_{y_{n}=1}^{M}\Pr ( y|x_{0}) ||x_{\perp }-x_{\perp }^{\prime
}( y) ||^{2}=\operatorname{constant}$.
\item To guarantee this constant, it is sufficient to have $\int
dx_{\perp } \Pr ( x_{\perp }|x_{0}) ||x_{\perp }-x_{\perp }^{\prime
}( y) ||^{2}=\operatorname{constant} \operatorname{independent}
\operatorname{of} x_{0} \operatorname{and} y$.
\item To guarantee this $\operatorname{constant} \operatorname{independent}
\operatorname{of} x_{0} \operatorname{and} y$, it is sufficient
to have $\Pr ( x_{\perp }|x_{0}) =\Pr ( x_{\perp }) $.
\item Given that $\Pr ( x_{\perp }|x_{0}) =\Pr ( x_{\perp }) $ and
$\Pr ( y|x_{0},x_{\perp }) =\Pr ( y|x_{0}) $, then making the replacement
$x_{\perp }^{\prime }( y) \longrightarrow <x_{\perp }>$ in $D$ will
give an objective function with the same stationary points as $D$,
because $x_{\perp }^{\prime }( y) =<x_{\perp }>$ is the stationary
point of $D$ with respect to $x_{\perp }^{\prime }( y) $.
\item Given that $\Pr ( x_{\perp }|x_{0}) =\Pr ( x_{\perp }) $ and
$x_{\perp }^{\prime }( y) =<x_{\perp }>$, the result $\int dx_{\perp
} \Pr ( x_{\perp }|x_{0}) ||x_{\perp }-x_{\perp }^{\prime }( y)
||^{2}=\operatorname{constant} \operatorname{independent} \operatorname{of}
x_{0} \operatorname{and} y$ follows automatically.
\end{enumerate}

\noindent The assumptions may be summarised as
\begin{equation}
\begin{array}{rl}
 \Pr ( y|x_{0},x_{\perp })  & =\Pr ( y|x_{0})  \\
 \Pr ( x_{\perp }|x_{0})  & =\Pr ( x_{\perp }) 
\end{array}%
\label{XRef-Equation-821215030}
\end{equation}

\noindent which allow the objective function $D$ (see Equation \ref{XRef-Equation-821214911})
to be replaced by the equivalent objective function
\begin{equation}
D=2\int dx_{0}\Pr ( x_{0}) \sum \limits_{y_{1}=1}^{M}\sum \limits_{y_{2}=1}^{M}\cdots
\sum \limits_{y_{n}=1}^{M}\Pr ( y|x_{0}) \left| \left| x_{0}-x_{0}^{\prime
}( y) \right| \right| ^{2}+\operatorname{constant}%
\label{XRef-Equation-821215049}
\end{equation}

\noindent This is the standard FMC objective function (compare Equation
\ref{XRef-Equation-821213533}) for encoding and reconstructing the
undistorted input, for which the directed graph is $\text{$x_{0}$}\longrightarrow
y\longrightarrow x_{0}^{\prime }$. Note that, under the stated assumptions,
the simplification in Equation \ref{XRef-Equation-821215049} occurs
even if the two subspaces are not orthogonal to each other, the
potential cross-term $\int dx_{\perp } \Pr ( x_{\perp }|x_{0}) (x_{0}-x_{0}^{\prime
}( y) ).(x_{\perp }-x_{\perp }^{\prime }( y) )$ in Equation \ref{XRef-Equation-821214911}
is zero.

In summary, the encoder has access only to the signal + distortion
$(x_{0},x_{\perp })$ (see Figure \ref{XRef-Figure-821214043} and
Equation \ref{XRef-Equation-821214911}), but the assumptions in
Equation \ref{XRef-Equation-821215030} force the encoder to disregard
the distortion (see Figure \ref{XRef-Figure-821214420} and Equation
\ref{XRef-Equation-821215049}). In practice, it is not possible
to satisfy these assumptions in general, because it is not known
in advance how to extract orthogonal signal and distortion subspaces
$(x_{0},x_{\perp })$ given examples of only the distorted signal.
However, these assumptions may be encouraged to hold true by minimising
$D$ (as defined in Equation \ref{XRef-Equation-821214911}) under
certain constraints, in which case Figure \ref{XRef-Figure-821214420}
and Equation \ref{XRef-Equation-821215049} follow automatically
from Figure \ref{XRef-Figure-821214043} and Equation \ref{XRef-Equation-821214911},
respectively. These constraints are discussed in Section \ref{XRef-Section-82121269}.

This type of encoder, in which the large degrees of freedom are
preferentially encoded, can be used as the basis of a so-called
"residual vector quantiser" \cite{BarnesRizviNasrabadi1996}, in
which (quoting from \cite{BarnesRizviNasrabadi1996}) "the quantiser
has a sequence of encoding stages, where each stage encodes the
residual (error) vector of the prior stage". Note that a residual
vector quantiser is a special case of the type of multistage encoder
discussed in \cite{Luttrell1999b}.
\subsection{Optimisation Constraints}\label{XRef-Section-82121269}

Henceforth, only the scalar case will be considered, so the vector
$y$ is now replaced by the scalar $y$ ($1\leq y\leq M$). In order
to implement a practical optimisation procedure for minimising $D$
it is necessary to introduce a variety of assumptions and constraints.

Because $\Pr ( y|x) $ is a probability it satisfies $\Pr ( y|x)
\geq 0$ and $\sum \limits_{y=1}^{M}\Pr ( y|x) =1$, which is guaranteed
if $\Pr ( y|x) $ is written as
\begin{equation}
\Pr ( y|x) =\frac{Q( y|x) }{\sum \limits_{y^{\prime }=1}^{M}Q( y^{\prime
}|x) }
\end{equation}

\noindent where $Q( y|x) \geq 0$. This removes the need to explicitly
impose the constraint $\sum \limits_{y=1}^{M}\Pr ( y|x) $ during
optimisation. The $Q( y|x) $ are the unnormalised {\itshape likelihoods}
of sampling code index $y$ from the code book.

However, $Q( y|x) $ itself needs to be described by a finite number
of parameters in order that the values that minimise $D_{1}+D_{2}$
may be derived from a finite amount of training data. It can be
shown that the optimal form of $\Pr ( y|x) $ is piecewise linear
in $x$ \cite{Luttrell1999b}, and that for training data that lie
on smooth curved manifolds the form of this solution is well approximated
by a piecewise linear $Q( y|x) $ of the form \cite{Luttrell1999a}
\begin{equation}
Q( y|x) =\left\{ \begin{array}{ccc}
 w( y) .x-a( y)  &   & w( y) .x\geq a( y)  \\
 0 &   & w( y) .x\leq a( y) 
\end{array}\right. 
\end{equation}

\noindent which is the same as the functional form used for the
neural response in \cite{Webber1994}. However, the precise functional
form of $Q( y|x) $ needs to exhibit this behaviour only in the vicinity
of the data manifold, so in particular it can be allowed to saturate
(i.e. $Q( y|x) \longrightarrow 1$) as $w( y) .x\longrightarrow \infty
$. A convenient functional form that achieves this is the sigmoid,
which is defined as
\begin{equation}
Q( y|x) =\frac{1}{1+\exp ( -w( y) .x-b( y) ) }
\end{equation}

\noindent This reduces the problem of minimising $D$ to one of finding
the optimal values of the $w( y) $, $b( y) $ and $x^{\prime }( y)
$. This may be done by using the gradient descent procedure described
in \cite{Luttrell1997}.

\begin{figure}[h]
\begin{center}
\includegraphics{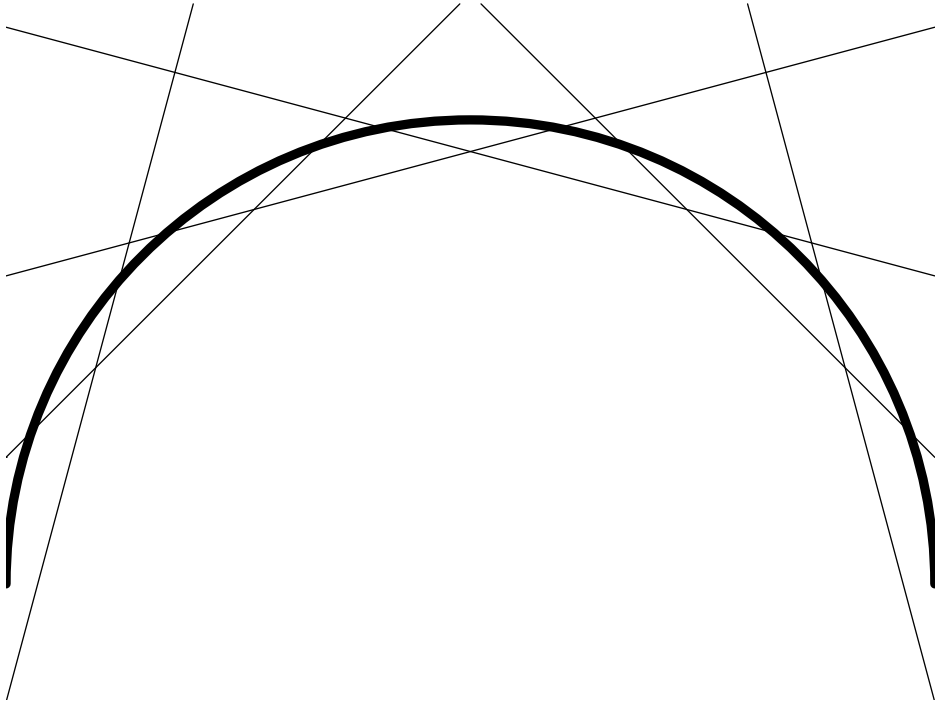}

\end{center}
\caption{Illustration of how a number of sigmoids can cooperate
to slice pieces off a signal manifold.}\label{XRef-Figure-82122212}
\end{figure}

If the input is an {\itshape undistorted} signal (i.e. $x=(x_{0},0)$)
which lies on a smooth curved manifold, then the sigmoids can cooperate
in encoding this input as illustrated in Figure \ref{XRef-Figure-82122212},
where the sigmoid threshold planes $w( y) .x+b( y) =0$ are shown
slicing pieces off the curved manifold \cite{Luttrell1999a}.

The additional constraints that are required in order to implement
the behaviour described in Section \ref{XRef-Section-821212555}
will now be described. Thus the constraints must be such that the
encoder disregards (or is invariant with respect to) the nuisance
degrees of freedom $x_{\perp }$ in the full input vector $(x_{0},x_{\perp
})$. However, without knowing $\Pr ( x_{0},x_{\perp }) $ in advance
(which would allow $x_{0}( x) $ in Equation \ref{XRef-Equation-821222147}
to be calculated), it is not possible to give a general approach
that works in all cases. At best, an empirical approach must be
used.

A very simple and useful constraint is to impose a threshold constraint
on the sigmoid function, which forces the value of the sigmoid to
lie exactly halfway up its slope when the norm of its input vector
is $\theta $. This is achieved by choosing $b( y) =-\theta |w( y)
|$, so that
\begin{equation}
Q( y|x) =\frac{1}{1+\exp ( -\left( \hat{w}( y) .x-\theta \right)
||w( y) ||) }%
\label{XRef-Equation-821223120}
\end{equation}

\noindent where $||w( y) ||\equiv \sqrt{w( y) .w( y) }$ and $\hat{w}(
y) \equiv \frac{w( y) }{||w( y) ||}$.

\begin{figure}[h]
\begin{center}
\includegraphics{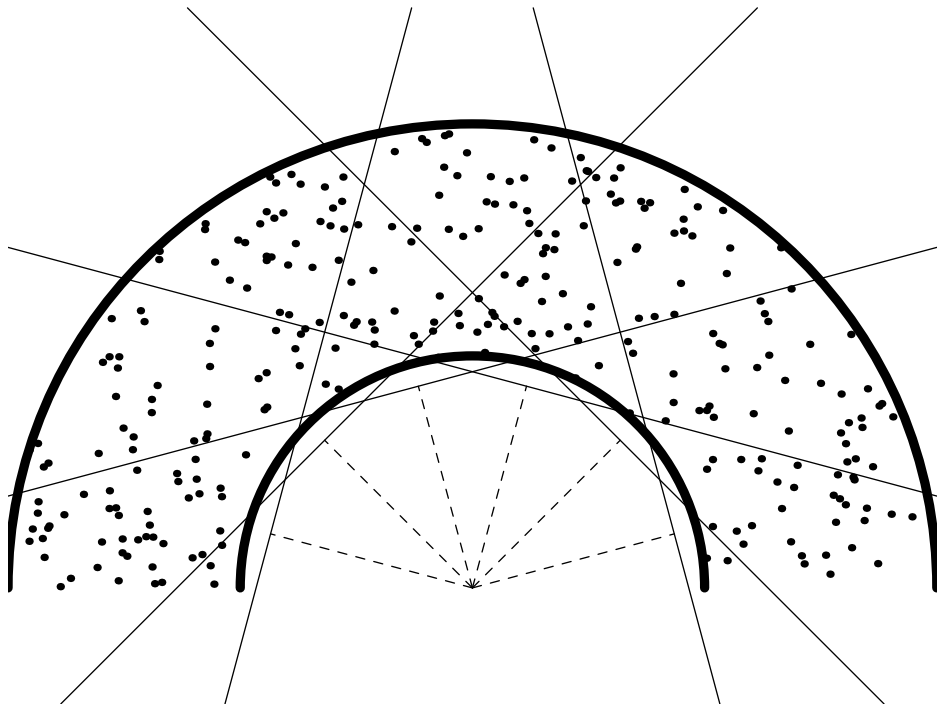}

\end{center}
\caption{Illustration of how a number of sigmoids can cooperate
to slice pieces off a signal manifold thickened by nuisance degrees
of freedom.}\label{XRef-Figure-82122230}
\end{figure}

If the input is a {\itshape distorted} signal (i.e. $x=(x_{0},x_{\perp
})$) which lies on a "thickened" version of the smooth curved manifold
of Figure \ref{XRef-Figure-82122212} (the thickness represents the
nuisance degrees of freedom), then the sigmoids can cooperate in
encoding this input as illustrated in Figure \ref{XRef-Figure-82122230},
where the sigmoid threshold planes $\hat{w}( y) .x=\theta $ are
shown slicing pieces off the curved manifold in a way that disregards
the nuisance degrees of freedom.
Note that in Figure \ref{XRef-Figure-82122230} the representation
of thickening is not complete, because it can actually occur in
{\itshape any} direction orthogonal to the manifold, including directions
orthogonal to the space in which the manifold is embedded; the radial
direction in Figure \ref{XRef-Figure-82122230} does not include
this latter possibility.

In practice, for numerical efficiency and to encourage the optimisation
procedure to locate the global minimum of $D_{1}+D_{2}$, it is useful
to introduce two additional constraints. Firstly, because optimal
solutions typically satisfy $x^{\prime }( y) \approx w( y) $ up
to a multiplicative constant, each reconstruction vector $x^{\prime
}( y) $ can be forced to lie parallel to the corresponding weight
vector $w( y) $, so that $x^{\prime }( y) \propto w( y) $; this
constraint was also used in \cite{Webber1998}, but there it was
a necessary part of the optimisation procedure, whereas here it
merely encourages faster convergence. Secondly, the norm of the
weight vectors $||w( y) ||$ can be constrained as $||w( y) ||=w_{0}$,
in order to avoid situations where they grow to rather large values
which make $Q( y|x) $ (and hence $\Pr ( y|x) $) depend very strongly
on $x$ in some regions. Both of these constraints speed up convergence
to the global minimum of\ \ $D_{1}+D_{2}$, can finally be lifted
in the vicinity of an optimal solution to obtain complete convergence.
\subsection{Jammer Nulling}\label{XRef-Section-821212651}

\begin{figure}[h]
\begin{center}
\includegraphics{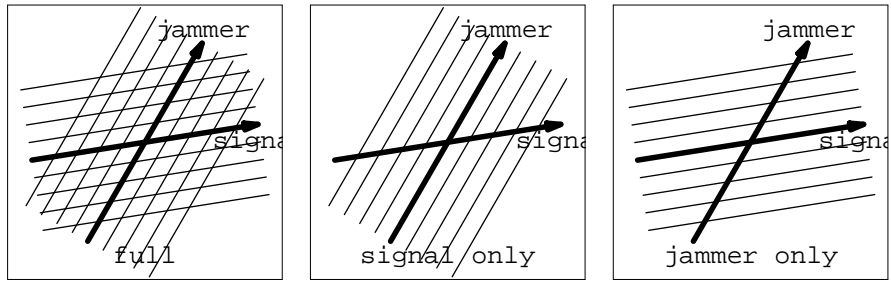}

\end{center}
\caption{Examples of the response of $\Pr ( y|x) $ to signal and
jammer subspaces.}\label{XRef-Figure-821222918}
\end{figure}

A number of examples of typical behaviours of $\Pr ( y|x) $ are
shown in Figure \ref{XRef-Figure-821222918}.

In Figure \ref{XRef-Figure-821222918} the signal and jammer degrees
of freedom generate a pair of non-orthogonal subspaces, whose axes
are indicated in bold. The response contours of a variety of possible
$\Pr ( y|x) $ are shown. In the "full" case a pair of $\Pr ( y|x)
$ respond to the signal and jammer subspaces respectively. In the
"signal" case a $\Pr ( y|x) $ responds to only the signal subspace,
and is thus invariant over the jammer subspace. In the "jammer"
case the situation is the reverse of the "signal" case. This argument
may readily be generalised to any number of $\Pr ( y|x) $.

If it is assumed that the jammer is the "large" degree of freedom
and the signal is the "small" degree of freedom, the signal and
jammer subspaces may be separated by adjusting the threshold parameter
$\theta $ so that in figure XXX the "jammer" case is obtained, in
which case the $\Pr ( y|x) $ for $y=1,2,\cdots ,M$ will all become
invariant over the signal subspace. The jammer subspace is then
spanned by the set of gradient vectors $\nabla \Pr ( y|x) $ for
$y=1,2,\cdots ,M$, which can thus be used to construct a projection
operator $J$ onto the jammer subspace, and a projection operator
$1-J$ onto the signal subspace. This definition of the projection
operator may also be used in cases where the jammer and signal subspaces
are curved, so that the directions of their axes are functions of
$x$, and all of the straight lines in Figure \ref{XRef-Figure-821222918}
are replaced by curves defining a curvilinear coordinate system
and its coordinate surfaces. Note that curved subspaces are the
norm rather than the exception.
\section{Jammer Nulling Simulations}\label{XRef-Section-82121246}

The optimisation of the encoder may be done by minimising $D$ using
gradient descent \cite{Luttrell1997}, using the sigmoid function
in Equation \ref{XRef-Equation-821223120} to constrain the optimisation
so that it encodes only the jammer subspace.

In these simulations the input vector $x$ is 100-dimensional so
that $x=(x_{1}, x_{2},\cdots ,x_{100})$, and each vector in the
training set is independently generated as a superposition of a
pair of response functions
\begin{equation}
x_{i}=a_{s} \frac{\sin ( \frac{i-i_{s}}{\sigma }) }{\frac{i-i_{s}}{\sigma
}}+a_{j}\frac{\sin ( \frac{i-i_{j}}{\sigma }) }{\frac{i-i_{j}}{\sigma
}}
\end{equation}

\noindent where $a_{s}$ is the signal amplitude that is uniformly
distributed in the interval $[-\sqrt{10^{-3}},\sqrt{10^{-3}}]$ (this
correponds to a signal level of -30dB), $a_{j}$ is the jammer amplitude
that is uniformly distributed in the interval $[-1,1]$ (this correponds
to a jammer level of 0dB), $i_{s}$ is the signal location that is
chosen to be 50, $i_{j}$ is the jammer location that is uniformly
distributed in the interval $[38-\Delta ,38+\Delta ]$ ($\Delta =0,2,4$
is used in the simulations), and $\sigma $ is the width of the response
function that is chosen to be 2. The peak and the first zero of
the sinc function are separated by $\pi  \sigma $, which defines
the resolution cell size. The mean jammer position and the signal
position satisfy $i_{s}-<i_{j}>=12$, which corresponds to a separation
of $\frac{12}{\pi  \sigma }\approx 2$ resolution cells. Random noise
uniformly distributed in the interval $[-\sqrt{10^{-5}},\sqrt{10^{-5}}]$
(this correponds to a noise level of -50dB) is also added to each
component of the training vector.
\begin{figure}[h]
\begin{center}
\includegraphics{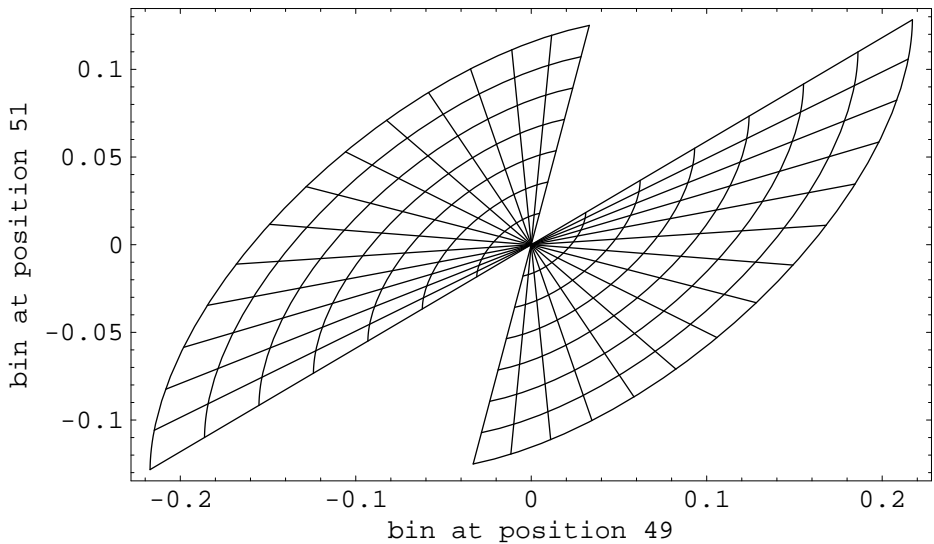}

\end{center}
\caption{A two-dimensional projection of the curved manifold generated
by the jammer when $\Delta =2$.}\label{XRef-Figure-821223428}
\end{figure}

In Figure \ref{XRef-Figure-821223428} the 2-dimensional manifold
generated by varying the jammer position over the interval $[38-\Delta
,38+\Delta ]$ (for $\Delta =2$), and varying the jammer amplitude
over the interval $[-1,1]$, is shown. Because the input vector $x$
is 100-dimensional, only a low-dimensional projection can be visualised,
and the 2-dimensional vector $(x_{49},x_{51})$ is displayed here.
The curvilinear grid traces out the coordinate surfaces of jammer
position $i_{j}$ and jammer amplitude $a_{j}$, and the whole diagram
shows how this grid is embedded in $(x_{49},x_{51})$-space. Note
that the $a_{j}$ dimension behaves as a "radial" coordinate (straight
lines), whereas the $i_{j}$ dimension behaves as an "angular" coordinate
(curved lines).
\begin{figure}[h]
\begin{center}
\includegraphics{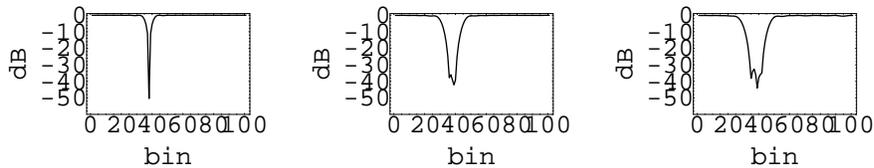}

\end{center}
\caption{Plot of degree of nulling against nominal jammer location,
for jammer locations that are spread over the intervals $[38,38]$
using $M=2$, $[36,40]$ using $M=4$, and $[34,42]$ using $M=6$.}\label{XRef-Figure-821223512}
\end{figure}

In Figure \ref{XRef-Figure-821223512} an encoder is trained on three
different jammer scenarios $\Delta =0,2,4$. After training the encoder
is tested for how well it can be used to null a pure jammer (i.e.
with no signal or noise added), where the degree of nulling is defined
as the ratio of the squared lengths of the nulled input vector and
the original input vector. This is a good test of the ability of
the encoder to simultaneously learn the profile of the jammer and
the shape of the jammer manifold which is generated by sweeping
this profile over the interval $[38-\Delta ,38+\Delta ]$. When $\Delta
=0$ there is a sharp minimum at the jammer location $i_{j}=38$,
as expected. When $\Delta =2$ the minimum becomes spread over the
jammer locations $i_{j}\in [36,40]$, and when $\Delta =4$ the minimum
becomes spread even more broadly over the jammer locations $i_{j}\in
[34,42]$. All of these results are as expected.
\begin{figure}[h]
\begin{center}
\includegraphics{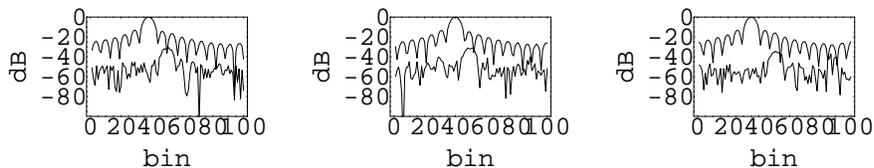}

\end{center}
\caption{Plot of a typical input vector before and after jammer
nulling for each of the scenarios in Figure \ref{XRef-Figure-821223512}.}\label{XRef-Figure-821223615}
\end{figure}

In Figure \ref{XRef-Figure-821223615} typical examples of an input
vector together with how it appears after jammer nulling are shown
for each of the jammer scenarios considered in Figure \ref{XRef-Figure-821223512}.
In every case the signal is clearly revealed at its correct location
after nulling the jammer.

In all of these training scenarios, one could envisage further constraining
some of the properties of the encoder, in order to introduce prior
knowledge of the form of the jammer and/or signal subspaces, and
to thereby reduce the computational complexity of the jammer nulling.
For instance, the signal subspace could be predefined, as in conventional
algorithms which hold constant the response in a predefined "look
direction". Similarly, the jammer subspace could be built out of
prefined subspaces which are optimised so as to maximally null the
jammer(s), as in conventional algorithms in which a number of jammer
"templates" are used to remove the jammer(s). In general, by choosing
appropriate additional constraints, the SVQ approach to jammer nulling
can be made backwardly compatible with conventional approaches.
\section{Conclusions}

The theory of stochastic vector quantisers (SVQ) \cite{Luttrell1997}
has been extended to allow the quantiser to develop invariances,
so that only "large" degrees of freedom in the input vector are
represented in the code. This has been applied to the problem of
encoding data vectors which are a superposition of a "large" jammer
and a "small" signal, so that only the jammer is represented in
the code. This allows the jammer to be subtracted from the total
input vector (i.e. the jammer is nulled), leaving a residual that
contains only the underlying signal. Several numerical simulations
have shown how that idea works in practice, even when the jammer
location is uncertain so that the jammer subspace is curved.

The main advantage of this approach to jammer nulling is that little
prior knowledge of the jammer is assumed, because these properties
are automatically discovered by the SVQ as it is trained on examples
of input vectors. Provided that the signal is much weaker than the
jammer, the SVQ acquires an internal representation of the jammer
and signal manifolds, in which its code is invariant with respect
to the signal. In a sense, the SVQ regards the "large" jammer as
the normal type of input that it expects to receive, whereas it
regards the "small" signal as an anomaly.

\appendix\label{TitleNote}

\end{document}